%% file: root.tex
\documentclass[10pt, journal, letterpaper,twoside]{IEEEtran} % Comment this line out if you need a4paper

\IEEEoverridecommandlockouts                              % This command is only needed if 
\makeatletter
\let\NAT@parse\undefined
\makeatother
\usepackage[numbers,sort&compress]{natbib}

\usepackage{comment}
\title{\LARGE \bf
Quadruped Obstacle Avoidance and Footstep Planning with Distributed Low-cost Time-of-Flight Sensors
}

\author{Giammarco Caroleo$^{1}$, Timoth\'ee Mahamoodally$^{1}$, Matteo Manzardo$^{2}$, Jin Jin$^{1}$, Marco Pontin$^{1}$,\\
Matias Mattamala$^{3}$, Renato Vidoni$^{2,4}$, Perla Maiolino$^{1}$, and Maurice Fallon$^{1}$% <-this % stops a space
\thanks{Manuscript received: May 4, 2026; Revised June 21, 2026; Accepted August 18, 2026.}
\thanks{This paper was recommended for publication by Editor Ayoung Kim upon evaluation of the Associate Editor and Reviewers' comments.}
\thanks{This work has been funded by the SESTOSENSO project (HORIZON EUROPE Research and Innovation Actions, GA number 101070310). For the purpose of open access, the authors have applied a Creative Commons Attribution (CC BY) license to any Accepted Manuscript version arising.}
\thanks{$^{1}$Oxford Robotics Institute, University of Oxford, UK. 
\texttt{\{giammarco, timoth\'ee, jinjin, mpontin, perla, mfallon\}@robots.ox.ac.uk}}%
\thanks{$^{2}$Free University of Bozen-Bolzano, Italy. \texttt{\{matteo.manzardo, renato.vidoni\}@unibz.it}}%
\thanks{$^{3}$University of Edinburgh, UK. \texttt{matias.mattamala@ed.ac.uk}}%
\thanks{$^{4}$University of Udine, Italy. \texttt{renato.vidoni@uniud.it}}
}
\usepackage{multicol}
\usepackage[pdfborder={0 0 0}]{hyperref}

\usepackage{graphicx}
\usepackage{amsmath,amsfonts,amssymb}
\usepackage[dvipsnames]{xcolor}
\usepackage{lipsum}
\usepackage{booktabs}
\usepackage{multirow}
\usepackage{siunitx}
\usepackage{cleveref}
\usepackage{duckuments}
\usepackage{subcaption}
\usepackage{hyperref}
\usepackage{caption}

\usepackage[acronym]{glossaries}
\glsdisablehyper

\newacronym{dof}{DoF}{Degree of Freedoms}
\newacronym{tof}{ToF}{Time-of-Flight}
\newacronym{fov}{FoV}{Field of View}
\newacronym{slam}{SLAM}{Simultaneous Localization and Mapping}
\newacronym{pc}{PC}{Point Cloud}

\newcommand{\anymal}{ANYmal D}

\newcommand{\rev}[1]{{\color{black} #1}}
\newcommand{\revv}[1]{{\color{black} #1}}
\newcommand{\shrink}[1]{{\color{black} #1}}

\def\secref#1{Sec.~\ref{#1}}

\def\figref#1{Fig.~\ref{#1}}

\def\eqref#1{Eq.~(\ref{#1})}

\begin{document}
% \markboth{IEEE ROBOTICS AND AUTOMATION LETTERS. PREPRINT VERSION. ACCEPTED SEPTEMBER, 2026}%
% {Caroleo \MakeLowercase{\emph{et al.}}: Quadruped Obstacle Avoidance with Distributed ToF Sensors}

\maketitle

\input{sections/00_abstract}
\input{sections/01_intro}

\input{sections/02_related_work}

\input{sections/03_system_design}

\input{sections/04_experiments}

\input{sections/05_conclusions}

% 
{\small
\bibliographystyle{IEEEtran}
\bibliography{references}
}

%                                   % the textheight of the last page by a suitable amount.
%                                   % This command does not take effect until the next page
%                                   % so it should come on the page before the last. Make
%                                   % sure that you do not shorten the textheight too much.

% %%%%%%%%%%%%%%%%%%%%%%%%%%%%%%%%%%%%%%%%%%%%%%%%%%%%%%%%%%%%%%%%%%%%%%%%%%%%%%%%
%\section*{APPENDIX}
%\section*{ACKNOWLEDGMENT}
%Acknowledgments - Removed for blind review

%%%%%%%%%%%%%%%%%%%%%%%%%%%%%%%%%%%%%%%%%%%%%%%%%%%%%%%%%%%%%%%%%%%%%%%%%%%%%%%%
\end{document}

%% file: sections/00_abstract.tex
\begin{abstract}
% Quadruped robots typically rely on depth cameras and LiDAR sensors to map their local environment. However, these sensors have limited close-range coverage, are relatively expensive, and consume significant power. This study investigates whether distributed Time-of-Flight (ToF) sensors can provide sufficient perception for common locomotion and navigation tasks.
% %
% We designed a distributed ToF sensing architecture for the ANYbotics ANYmal quadruped and benchmarked it against LiDAR and depth cameras for the tasks of terrain mapping and obstacle avoidance. Distributing these sensors around the robot can also avoid the blind spots of traditional sensors. Our results show that, despite their low resolution and higher measurement noise, distributed ToF sensors can support reliable perceptual locomotion with centimeter-level local mapping accuracy. The proposed sensing strategy \rev{provides sufficient geometric information for near-field obstacle avoidance and footstep planning, offering a competitive alternative to depth cameras in terms of cost, energy consumption, and system complexity.}
% The proposed sensing strategy achieves comparable performance to the typical solution based on depth cameras while offering a competitive alternative in terms of cost, energy consumption, and system complexity.
Quadruped robots typically rely on depth cameras and LiDAR sensors to map their local environment. However, these sensors have limited close-range coverage, are relatively expensive, and consume significant power. This study investigates whether distributed Time-of-Flight (ToF) sensors can \revv{serve as a low-cost alternative to depth cameras for near-field terrain mapping for locomotion and local navigation}.
We designed a distributed ToF sensing architecture for the ANYbotics ANYmal quadruped\revv{, assessed its environment reconstruction accuracy, and benchmarked it against depth cameras for terrain mapping and obstacle avoidance}. Distributing these sensors around the robot can also avoid the blind spots of traditional sensors. Our results show that, despite their low resolution and higher measurement noise, distributed ToF sensors can support reliable perceptual locomotion with centimeter-level local mapping accuracy. The proposed sensing strategy \rev{provides sufficient geometric information for near-field obstacle avoidance and footstep planning}\revv{, at substantially lower cost, energy consumption, and system complexity than depth cameras.}
\end{abstract}

%% file: sections/01_intro.tex
\section{Introduction}

Quadruped robots have been demonstrated in a wide range of applications, including inspection \cite{Bellicoso2018}, industrial monitoring \cite{staniaszek2025autoinspect}, and forest mapping in complex and unstructured environments \cite{mattamala2025tfr}. 
Their ability to traverse uneven terrain and navigate obstacles \revv{suits these challenging scenarios, where robust autonomy depends on reliable perception and environment understanding.}
% makes them particularly suitable for these challenging scenarios. However, achieving robust autonomy in such settings depends on reliable perception and environment understanding.

Most state-of-the-art quadruped platforms rely on information-rich sensing modalities such as depth cameras and LiDAR for perception, mapping, and navigation~\cite{kim2020icra, ramezani2020icra}. While these sensors enable accurate state estimation and dense environment reconstruction, they have some important limitations. 
Depth cameras are rather large (the RealSense D435i is 9~cm wide) and cannot be placed on curved surfaces or close to joints, \revv{resulting} in blind spots around the legs and feet -- critical locations for safe foothold selection. LiDAR provides accurate long-range measurements but is less effective at short range and may fail to capture fine geometric details required for local terrain interaction. Both modalities are also relatively expensive and have high computational and power requirements.

% \mfallon{"Cameras are sensitive to illumination changes" This is not true - depth cameras are active sensors with IR illumination and are not lighting sensitive}

Recent advances in multizone \gls{tof} sensors, \revv{which} provide multiple distance measurements within their \gls{fov},  have enabled distributed, low-cost perception~\cite{adamides2019, kumar2019smc, caroleo2025iros, teetaert2026continuum}. While individual sensors provide low-resolution, noisy measurements~\cite{caroleo2026characterisation}, fusing the information provided by 
% a set of 
distributed \gls{tof} sensors increases coverage and robustness. At present, their suitability for legged robot navigation and locomotion has been largely unexplored.

\begin{figure}[t]
    \centering
    \captionsetup{font=small}
    \includegraphics[width=0.9\linewidth]{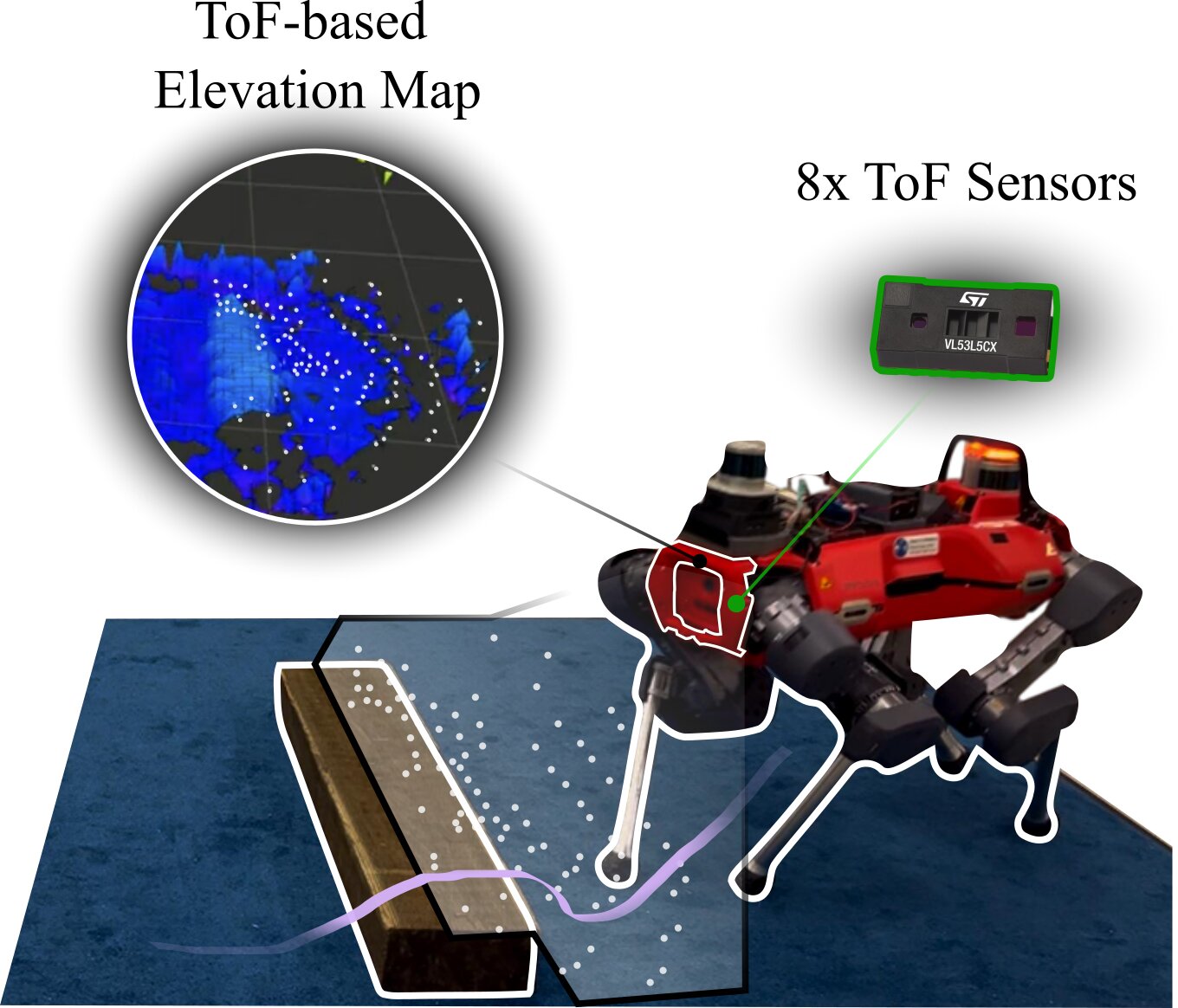}
\caption{This paper studies time-of-flight (ToF) sensors for quadrupedal locomotion and navigation. Mounted on the front shell of ANYmal D, they provide near-field perception for elevation mapping, obstacle avoidance, and footstep planning.}
% This paper studies the use of time-of-flight (ToF) sensors for quadrupedal locomotion and navigation. Multiple ToF sensors\shrink{, mounted on the front shell of \anymal{},}
% are mounted on the front shell of \anymal{} to
% provide near-field perception for elevation map generation, obstacle avoidance and footstep planning.}
\label{fig:intro}
\end{figure}

In this article, we investigate whether a distributed array of off-the-shelf \gls{tof} devices can serve as \revv{a low-cost alternative to depth cameras for near-field} \rev{terrain perception} (Fig.~\ref{fig:intro})\revv{, providing a design exploration of low-cost perception for legged robots.}
% . Our study provides a benchmarking and design exploration of low-cost perception for legged robots.

We \revv{compare it against depth cameras for terrain mapping, navigation, and footstep planning, and assess its geometric reconstruction accuracy against a LiDAR-based reference}.

The contributions of our study are:
\begin{enumerate}
    \item We \revv{evaluate the geometric reconstruction accuracy of distributed \gls{tof} sensors against a conventional LiDAR reference (Hesai Pandar QT64)}, highlighting trade-offs in accuracy, coverage, and data efficiency.
    \item We propose an integration design of distributed \gls{tof} sensors on the front shell of the ANYbotics \anymal{}, enabling perception of regions typically occluded to standard sensing setups, e.g. near the robot’s feet. 
    \item \rev{We validate the sensing system in simulated and real-world experiments, demonstrating that distributed ToF sensing provides sufficient near-field terrain perception for obstacle avoidance and footstep planning in cluttered environments}\revv{, at a fraction of the cost of depth cameras.}
\end{enumerate}

%% file: sections/02_related_work.tex
\section{Related Work}
\label{sec:literature}

\subsection{Geometric Perception for Quadrupeds}
Geometric sensing modalities, such as depth cameras and LiDAR, are the preferred sensing modality for core quadrupedal locomotion capabilities, such as footstep planning, local obstacle avoidance, and long-range path planning. 

LiDAR provides long-range, accurate, and relatively sparse depth measurements, which has been used primarily for state estimation and mapping \cite{ou2024, wisth2022vilens} in legged robots. This has enabled different field applications of legged robots, including underground exploration~\cite{bouman2020autonomous, tranzato2024cerberus, agha2021nebula}, forestry \cite{mattamala2025tfr} and prospective planetary missions \cite{ arm2023scientific}.

Depth cameras, conversely, provide dense but close-range sensing, being more suitable for local terrain mapping. These local maps are usually used for locomotion with model-based~\cite{grandia2023tro, Kolter09} or learning-based approaches \cite{miki2022learning}. Furthermore, they have also enabled downstream tasks such as navigation \cite{buchanan2019walking, zhang2024resilient, wellhausen2023artplanner}. Recent works have also demonstrated the effectiveness of passing raw depth input directly to the quadruped's locomotion or navigation systems. One of the most recent examples is a robot parkour system, enabling direct locomotion control from single depth setups \cite{cheng2024extreme} or omnidirectional depth sensing using multiple cameras \cite{hoeller2024scirob}.

In this work, we explore the use of distributed ToF sensors as an alternative close-range geometric sensor to depth cameras. We demonstrate their potential in quadruped-relevant tasks, where depth cameras are currently preferred, such as foothold planning and local obstacle avoidance.

\subsection{Close-range Perception}
Close-range perception provides local awareness capabilities to mobile robots, which might be beneficial for reactive behaviors and collision avoidance.
% miniature \gls{tof} sensors have been widely investigated for close-range perception tasks due to their compact size, low power consumption, and low cost. Their application spans from obstacle detection and human-robot interaction in collaborative robotics~\cite{kumar2019case, ding2019iros, caroleo2024iros, borelli2025}, to state estimation in deformable and continuum robots~\cite{abah2022, caroleo2025robosoft, teetaert2026continuum}.
Early works explored single-beam sonar and ultrasonic sensors~\cite{elfes1987sonar, borenstein2002obstacle}, both for reactive obstacle avoidance and occupancy-based mapping, but relied on the assumption that the robot moves on flat ground. Biologically-inspired alternatives such as artificial whiskers~\cite{yu2022whisker}, instead, enable navigation without such assumptions by sensing local geometry around the robot body; however, in this context, knowledge is acquired only upon near-contact. 

% Their use in mobile robotics is more limited, with notable applications in nano-UAVs, where obstacle avoidance with ToF measurements is attractive given the strict payload and energy constraints~\cite{benini2023, benini2024}. In these settings, measurements are often projected onto simplified polar/range representations, as only coarse environmental awareness is required.
More recently, \gls{tof} sensors have emerged as a lightweight alternative for obstacle avoidance on nano-UAVs~\cite{decroon2022insect}, where using depth measurements from \gls{tof} is attractive given the strict payload and energy constraints~\cite{fuller2023gyroscope, benini2024}. In these settings, measurements are often projected onto simplified polar/range representations, as only coarse environmental awareness is required. Their use for perception, control, and 3D reconstruction has been more widely explored in robotic manipulators~\cite{caroleo2024iros, borelli2025} and continuum robots~\cite{abah2022, teetaert2026continuum}, where spatial structure around the robot body is needed for proximity-aware control.

Extending such approaches to quadruped robots introduces additional challenges. Legged locomotion requires spatially structured terrain information to support elevation mapping and safe foothold selection, particularly in close proximity to the robot. In this context, it remains underexplored whether distributed low-resolution ToF sensing can achieve sufficiently accurate reconstruction to support \rev{local} navigation and locomotion tasks, which is the gap we address in this work.

%% file: sections/03_system_design.tex
\section{System Design}\label{sec:system_design}
In this section we present the design choices made to physically integrate the sensors on the quadruped robot, along with the software integration with the navigation and locomotion stack. 

\subsection{ToF Sensors}
We used the VL53L5CX ToF sensors from STMicroelectronics\footnote{VL53L5CX Time-of-Flight (ToF) 8x8 multizone ranging sensor with wide field of view. [Online] \url{https://www.st.com/en/imaging-and-photonics-solutions/vl53l5cx.html}}.
The individual VL53L5CX module measures $6.4 \times 3.0 \times 1.5$ \si{mm} and 
produce an $8\times8$ multizone depth image with a $65$\si{\degree} diagonal \gls{fov} in a range between 0.02\,m and 4\,m. These measurements can be converted into point clouds for coarse sense reconstruction. 

From a cost perspective, a single VL53L5CX sensor costs approximately USD\$10, while a RealSense D435i depth camera costs around USD\$400. Achieving comparable field-of-view coverage with multiple ToF sensors therefore results in an order-of-magnitude reduction in sensing cost. 

In terms of energy consumption, each ToF sensor module draws approximately 0.2--0.4~W, compared to 2--4~W for a depth camera, leading to significantly lower power requirements for equivalent coverage.

\begin{figure}[t]
    \centering
    \captionsetup{font=small}
    \includegraphics[width=0.9\linewidth]{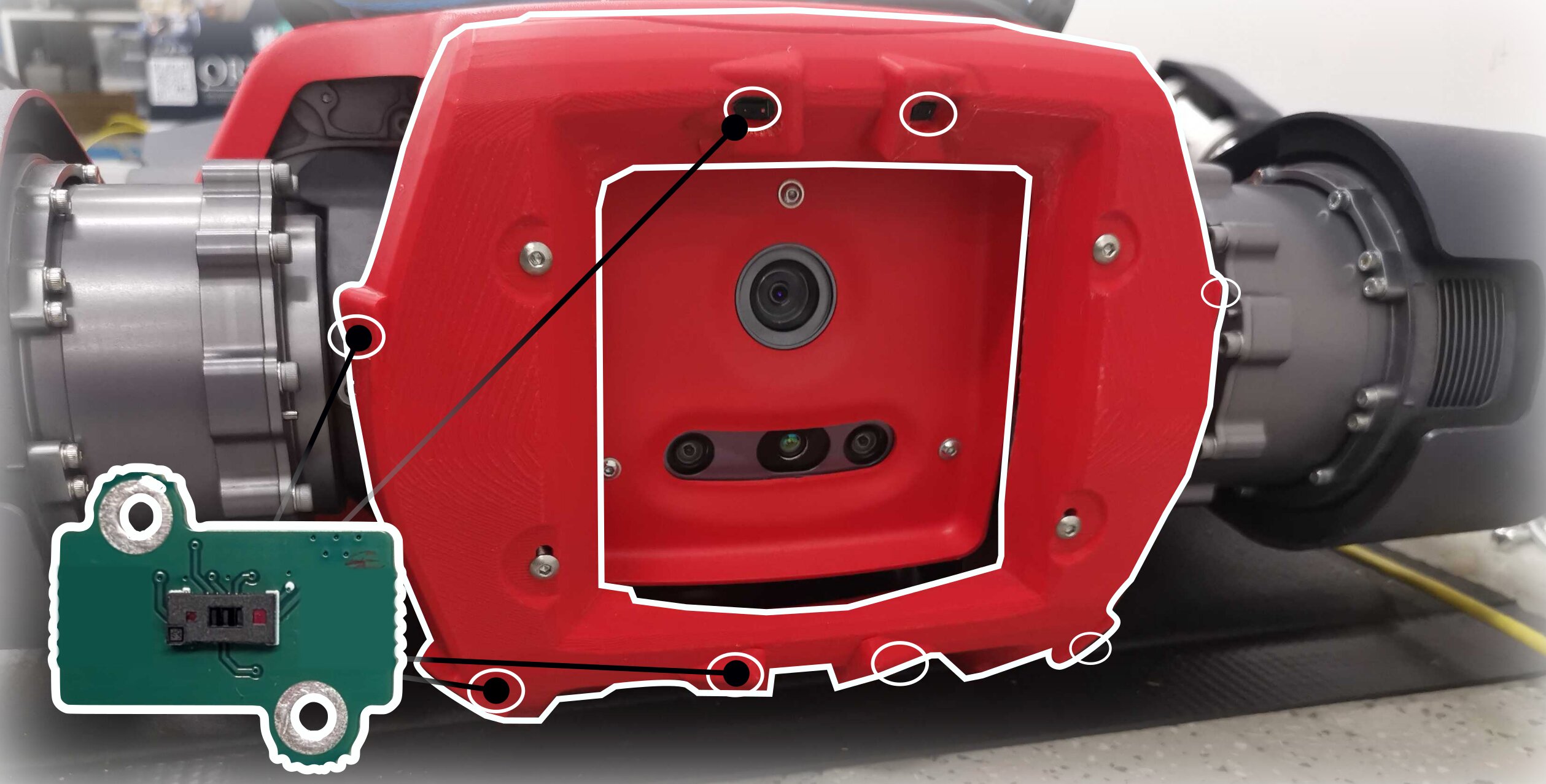}
    \caption{The 3D-printed front shell embedding the \gls{tof} sensors. The shell integrating 8 VL53L5CX is highlighted. The inset view shows the custom PCB designed for the sensing unit.}
    \label{fig:custom_shell}
\end{figure}

% \begin{figure*}[!t]
% \includegraphics[width=0.95\textwidth]{figures/Artboard 2.png}
% \caption{Field of View of the ANYmal's ToF sensors and depth cameras. (a) Shaded pyramids representing the FoV of each ToF sensor. (b) Angular (azimuth–elevation) coverage of the eight ToF sensors compared against the two onboard depth cameras. (c) FoV of the depth cameras and the distributed ToFs projected on the ground plane.}
% \label{fig:fov_image}
% \end{figure*}
%\begin{figure*}[t]
%    \centering
%    \captionsetup{font=small}
%    \begin{subfigure}[b]{0.31\textwidth}
%        \centering

%\includegraphics[width=0.9\linewidth]{figures/anymal_d.jpg}                \caption{Shaded pyramids representing the \gls{fov} of each \gls{tof}.}
%\label{fig:fov_a}
%    \end{subfigure}
%    \hfill
%    \begin{subfigure}[b]{0.31\textwidth}
%        \centering
%        \includegraphics[width=0.9\linewidth]{figures/azimuth.jpg}              \caption{Azimuth--elevation coverage of the eight \gls{tof}s compared against the depth cameras.}

 %       \label{fig:fov_b}
 %   \end{subfigure}
 %   \hfill
 %   \begin{subfigure}[b]{0.31\textwidth}
 %       \centering
 %       \includegraphics[width=\linewidth]{figures/fov_anymal.jpg}
 %        \caption{The FoV of the depth cameras and the ToFs when projected on the ground plane.}\label{fig:fov_c}
 %   \end{subfigure}
 %   \caption{Field of View of the ANYmal's \gls{tof} sensors (a), compared against its depth cameras in angular (b) and ground-plane (c) coverage.}
%\label{fig:fov_image}
%\end{figure*}

\begin{figure*}[t]
\centering
\captionsetup{font=small}
\includegraphics[width=0.9\linewidth]{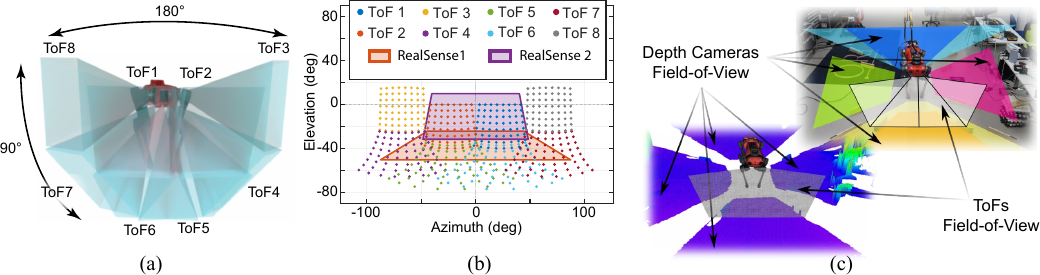}  
\caption{Field of View of the ANYmal’s \gls{tof} sensors. (a) Shaded pyramids representing the \gls{fov} of each \gls{tof}. (b) Azimuth–elevation coverage of the eight \gls{tof}s compared against the depth cameras. (c) The \gls{fov} of the depth cameras and the \gls{tof}s when projected on the ground plane.}
\label{fig:fov_image}
\end{figure*}

\subsection{ANYmal Integration}
We customized the front shell design of \anymal{} to host multiple sensors as shown in \figref{fig:custom_shell}. The \gls{tof} sensors are spaced to cover the region immediately in front of the quadruped to detect obstacles at close range and support reactive navigation. We heuristically determined the number and placement of sensing units to maximize coverage of the frontal workspace while maintaining mechanical feasibility. 

The adopted sensor arrangement introduces partial overlap between neighboring \gls{fov}s (see \figref{fig:fov_image}). \shrink{This increases the spatial density of near-field depth measurements, improving reliability under noise and partial occlusion.}
% This design choice increases the spatial density of depth measurements in the near field, improving reliability in the presence of noise and partial occlusions. 
At the same time, the placement was \rev{chosen} to minimize self-observations, i.e., to avoid large portions of the robot’s body falling within the sensing \gls{fov}s and corrupting the measurements.

A custom PCB was designed to rigidly mount the sensing units onto the 3D-printed robot’s front shell using screw fixtures, as detailed in the inset image in \figref{fig:custom_shell}. \rev{The final configuration  spans $180$\si{\degree} horizontally and $90$\si{\degree} vertically, as shown in \figref{fig:fov_image}a, thus covering some of the blindspots of the distributed depth cameras on ANYmal D (87°×58° FoV per camera, shown in \figref{fig:fov_image}c). \figref{fig:fov_image}b compares the resulting angular coverage of the ToF array against the two onboard depth cameras, showing extended azimuthal and elevation coverage beyond the cameras' field of view.}

\subsection{Software Integration}
\begin{figure}
    \centering
     \captionsetup{font=small}
    \begin{subfigure}[t]{0.48\linewidth}
        \centering
        \includegraphics[width=\linewidth]{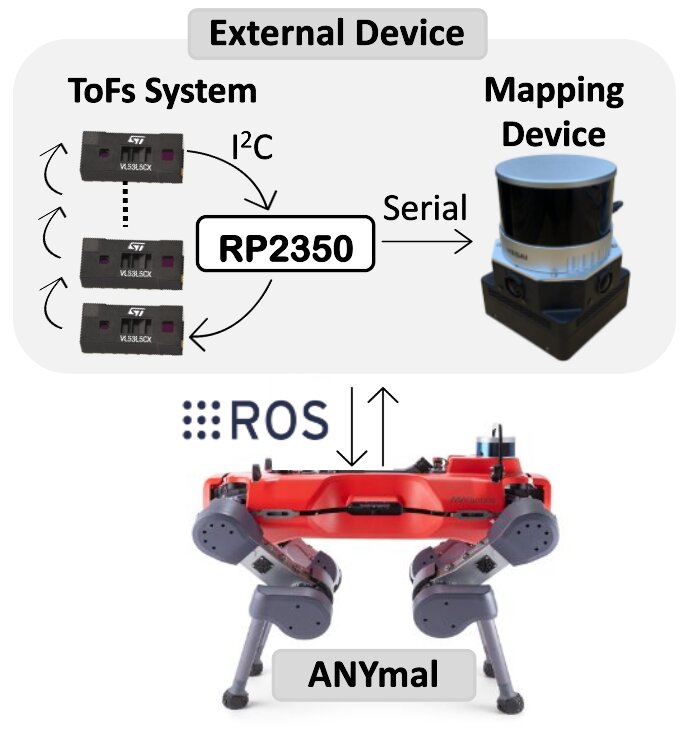}
        \caption{Integration of the ToF-based perception system within \anymal's architecture.}
        \label{fig:designdiagram}
    \end{subfigure}
    \hfill
    \begin{subfigure}[t]{0.48\linewidth}
        \centering
        \includegraphics[width=\linewidth,height=1.6in,keepaspectratio]{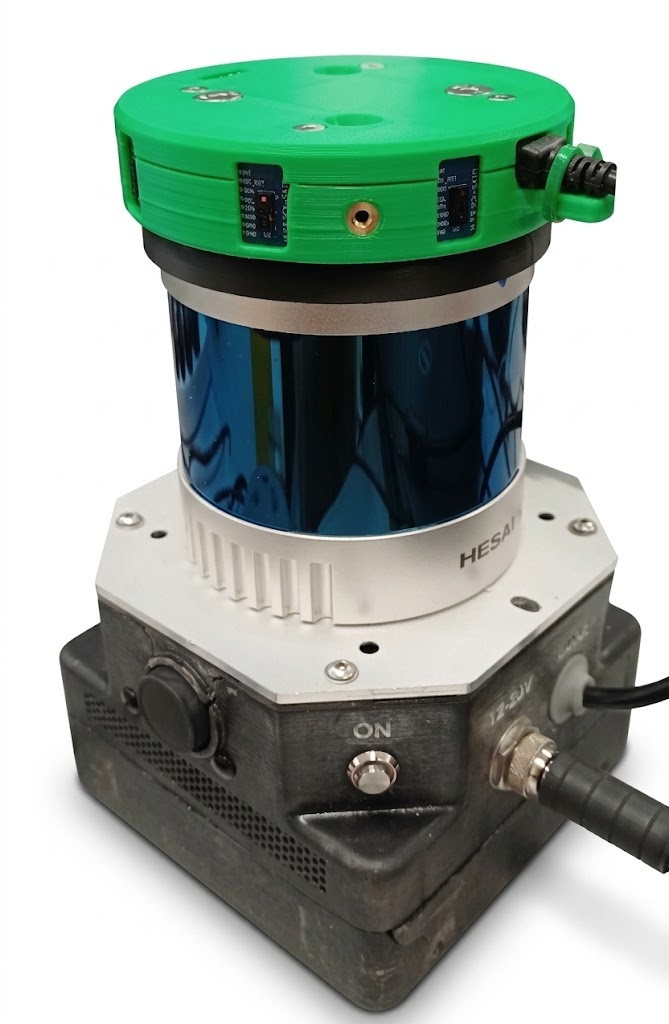}
        \caption{LiDAR mapping device with a ring of 8 equally spaced ToF sensors used for handheld reconstruction experiments in \cref{ssec:reconstruction}.}
        \label{fig:handheld_experiments}
    \end{subfigure}
    \caption{System-level integration of ToFs. We show the \anymal{} and handheld reconstruction setup.}
    \label{fig:system_and_handheld}
\end{figure}

Communication with the ToF sensors is daisy-chained in sequence (\figref{fig:designdiagram}), using the  I\textsuperscript{2}C protocol, \rev{with readout across all eight sensors completing in approximately 1 ms per cycle,} and the data are sampled at \SI{15}{\hertz}. Sensors are connected to a custom read-out board using an RP2350 microcontroller\footnote{RP2350 Datasheet. [Online] \url{https://www.alldatasheet.com/datasheet-pdf/pdf/2189433/RASPBERRY-PI/RP2350.html}} connected via USB to the communication port of the LiDAR mapping device, which was then mounted on top of \anymal{} and supplied power.
\shrink{A custom driver streamed ToF point clouds in ROS2, with extrinsic calibration obtained from the CAD model used to house the ToFs on ANYmal D's front face.}
% A custom driver was used to stream \gls{tof} point cloud messages in ROS2. 
% The extrinsic calibration of the sensor was obtained from the CAD model designed to house the \gls{tof}s in \anymal{}'s front face.

The point cloud measurements were then integrated within the robot's footstep planning and the navigation pipelines through the elevation mapping framework by Fankhauser et al. \cite{fankhauser2018ral}\rev{, running at 40 Hz}. 
This required projecting \gls{tof} data into the robot-centric map frame by considering the noise associated with them. We adapted the sensor model proposed by Nguyen et al. \cite{nguyen2012} by tuning the normal factors {\it a}, {\it b}, {\it c}, and {\it d} to $0.01, 0.02, 2$, and $1$, and the lateral one to $0.014$, to describe the ToFs. 
\rev{The robot's pose was provided by ANYmal D's onboard multi-sensor state estimator, fusing proprioceptive sensing with onboard LiDAR odometry and was used for both configurations. Neither the ToF sensors nor the depth cameras contribute to localization, as they are used exclusively to build the elevation map.}

\subsection{Handheld Mapping Device Integration}
\label{ssec:handheld-integration}

% Lastly, to enable comparison in reconstruction tasks, we also integrated the same ToF sensors into a separate handheld LiDAR mapping device, shown in  \figref{fig:handheld_experiments}.
\shrink{To enable comparison in reconstruction tasks, we also integrated the same ToF sensors into a separate handheld LiDAR mapping device (\figref{fig:handheld_experiments}).}
The mapping device carries a Hesai PandarQT-64, which provides $360$\si{\degree} horizontal coverage and $\approx104$\si{\degree} vertical coverage.

We mounted 8 ToF sensors in the 3D-printed shell attached to the top of the LiDAR unit. The ToF modules were arranged in a uniform ring to provide $360$\si{\degree} horizontal coverage and $45$\si{\degree} vertical coverage. \rev{This arrangement consists of a simpler sensor configuration than the one designed for \anymal's front shell, and targeted full horizontal coverage for handheld scanning rather than matching the LiDAR's vertical field of view.} The mapping device runs ROS2, drivers, and further processing onboard.

% The final configuration spans $180$\si{\degree} horizontally and $90$\si{\degree} vertically, as shown in \figref{fig:fovs_a}, thus covering some of the blindspots of the distributed depth cameras on \anymal{} \rev{(87° × 58° FoV per camera, }\figref{fig:fovs_b}).

% \begin{figure}[!t]
%     \centering
%     \captionsetup{font=small}
%      \begin{subfigure}[b]{0.4\textwidth}
%          \centering
%          \includegraphics[width=0.8\linewidth]{figures/anymal_d.jpg}
%          \caption{Shaded pyramids representing the \gls{fov} of each \gls{tof}.}
%          \label{fig:fovs_a}
%      \end{subfigure}
%      \hfill
%      \begin{subfigure}[b]{0.5\textwidth}
%          \centering
%          \includegraphics[width=\linewidth]{figures/fov_anymal.jpg}
%          \caption{The FoV of the depth cameras and the distributed ToFs when projected on the ground plane.}
%          \label{fig:fovs_b}
%      \end{subfigure}
%     % \includegraphics[width=\linewidth]{figures/anymal_d.png}
%     \caption{Field of View of the ANYmal's ToF sensors (a) and its standard depth cameras (b).
%     }
%     \label{fig:fov_image}
% \end{figure}

%% file: sections/04_experiments.tex
\section{Experimental Evaluation}\label{sec:experiments}

\begin{figure}[!t]
    \centering
    \captionsetup{font=small}
    \begin{subfigure}[t]{0.49\linewidth}
        \centering
        \includegraphics[width=\linewidth,height=2.1in,keepaspectratio]{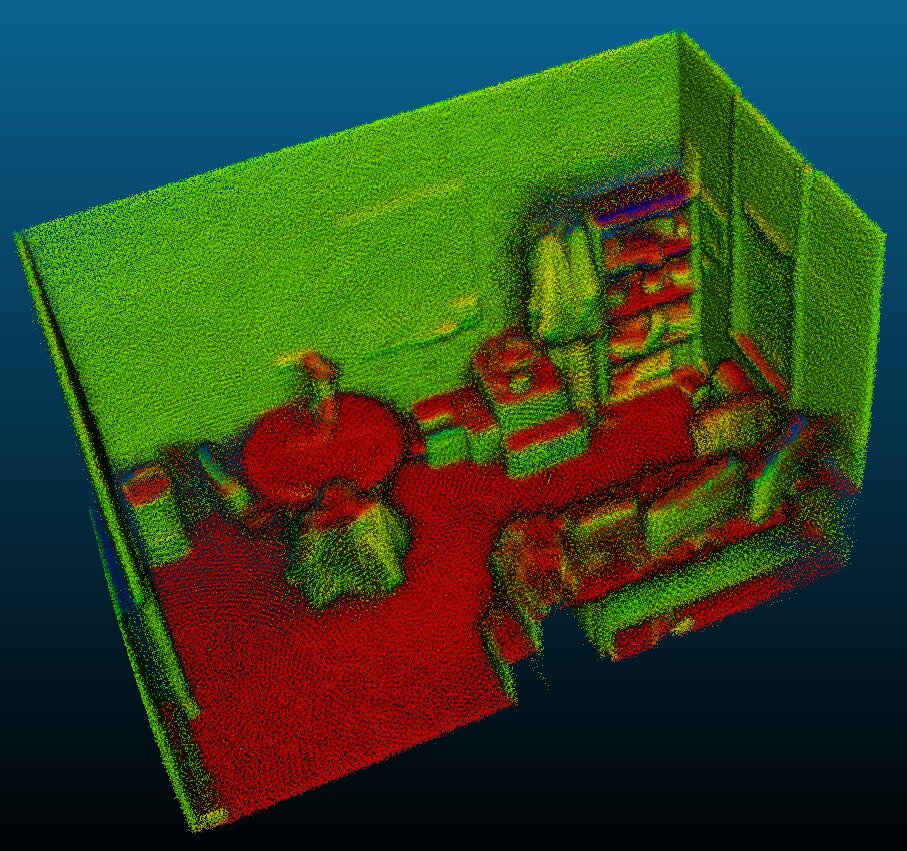}
    \end{subfigure}
    \hfill
    \begin{subfigure}[t]{0.49\linewidth}
        \centering
        \includegraphics[width=\linewidth,height=2.1in,keepaspectratio]{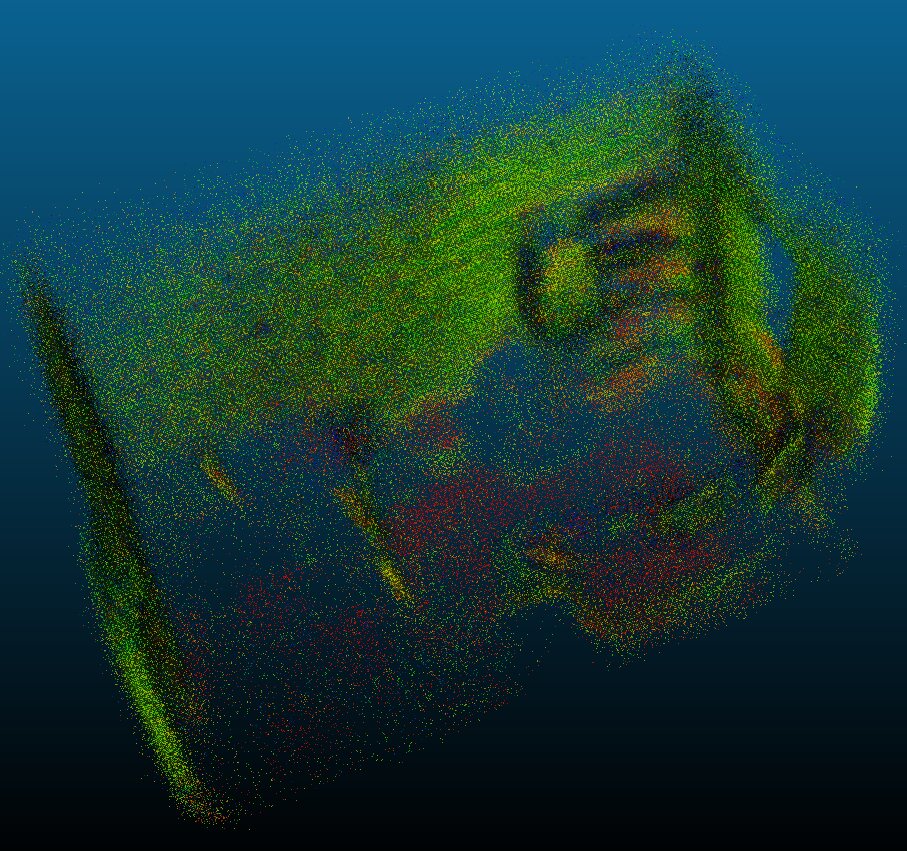}
    \end{subfigure}
    
    \caption{Accumulated point cloud reconstruction results of an office using LiDAR (left) and ToF (right). The point clouds are colorized by surface normal in the z-direction.}
    \label{fig:pc_comparison_views}

    \vspace{0.5em}
    \centering
    \captionof{table}{Per-point distance metrics between ToF and LiDAR point clouds in both directions.}
    \label{tab:tof_lidar_distance_metrics}

    % We make the font of the table smaller
    \footnotesize
\begin{tabular}{lcc}
\toprule
\multirow{2}{*}{\textbf{Metric}} & \multicolumn{2}{c}{\textbf{Distance Direction}} \\
\cmidrule(lr){2-3}
 & \textbf{ToF$\rightarrow$LiDAR} & \textbf{LiDAR$\rightarrow$ToF} \\
\midrule
Mean (m)   & 0.043 & 0.038 \\
RMSE (m)   & 0.074 & 0.067 \\
Median (m) & 0.019 & 0.017 \\
P90 (m)    & 0.108 & 0.105 \\
P95 (m)    & 0.174 & 0.159 \\
Max (m)    & 0.834 & 0.760 \\
\bottomrule
\end{tabular}
\end{figure}

\begin{figure}[!t]
    \centering
    \captionsetup{font=small}
    \begin{subfigure}[t]{0.49\linewidth}
        \centering
        \includegraphics[width=\linewidth,height=2.1in,keepaspectratio]{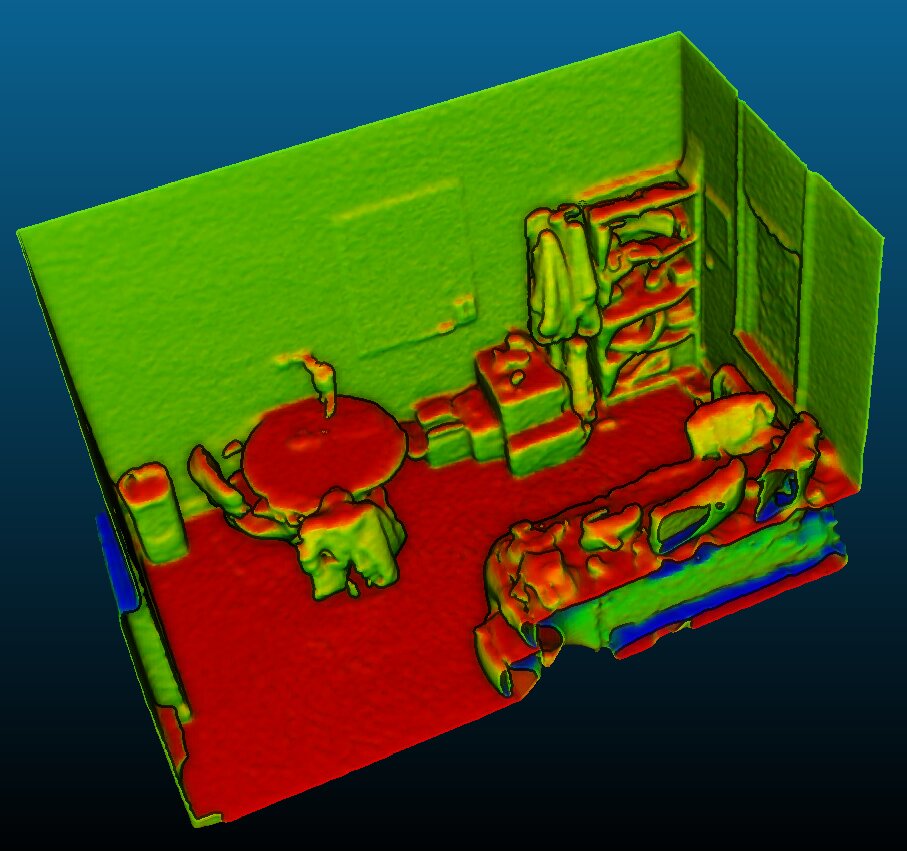}
        \caption{LiDAR Poisson mesh}
        \label{fig:lidar_poisson_reference}
    \end{subfigure}
    \hfill
    \begin{subfigure}[t]{0.49\linewidth}
        \centering
        \includegraphics[width=\linewidth,height=2.1in,keepaspectratio]{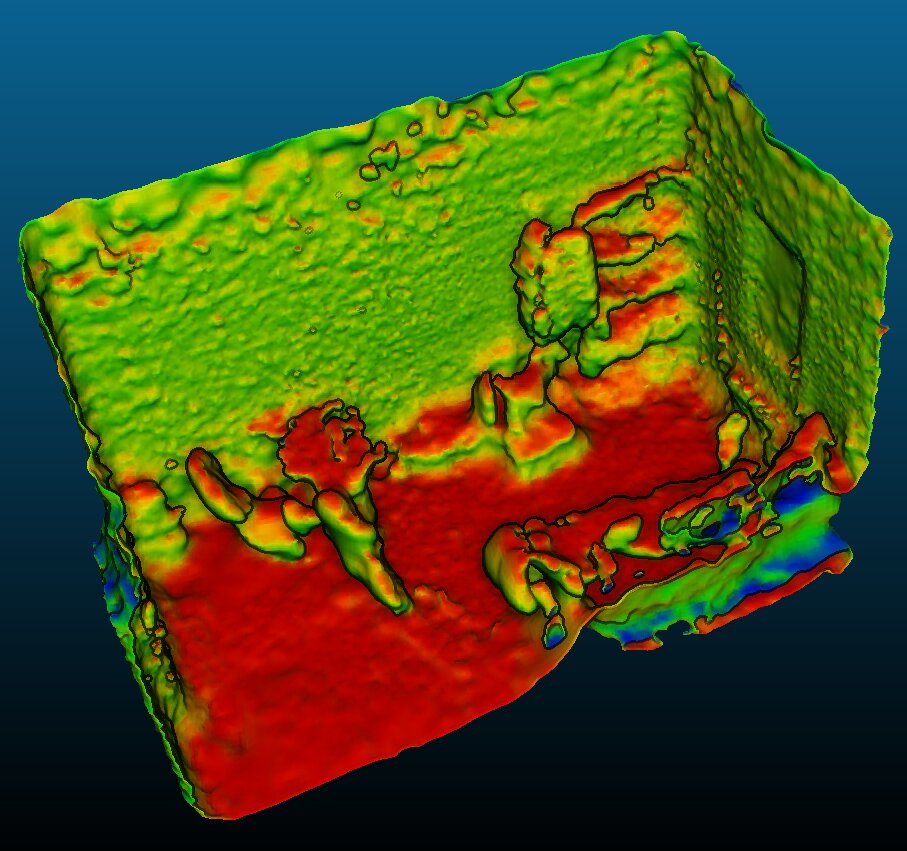}
        \caption{ToF Poisson mesh}
        \label{fig:tof_poisson_a}
    \end{subfigure}

    \vspace{0.5em}

    \begin{subfigure}[t]{0.49\linewidth}
        \centering
        \includegraphics[width=\linewidth,height=2.1in,keepaspectratio]{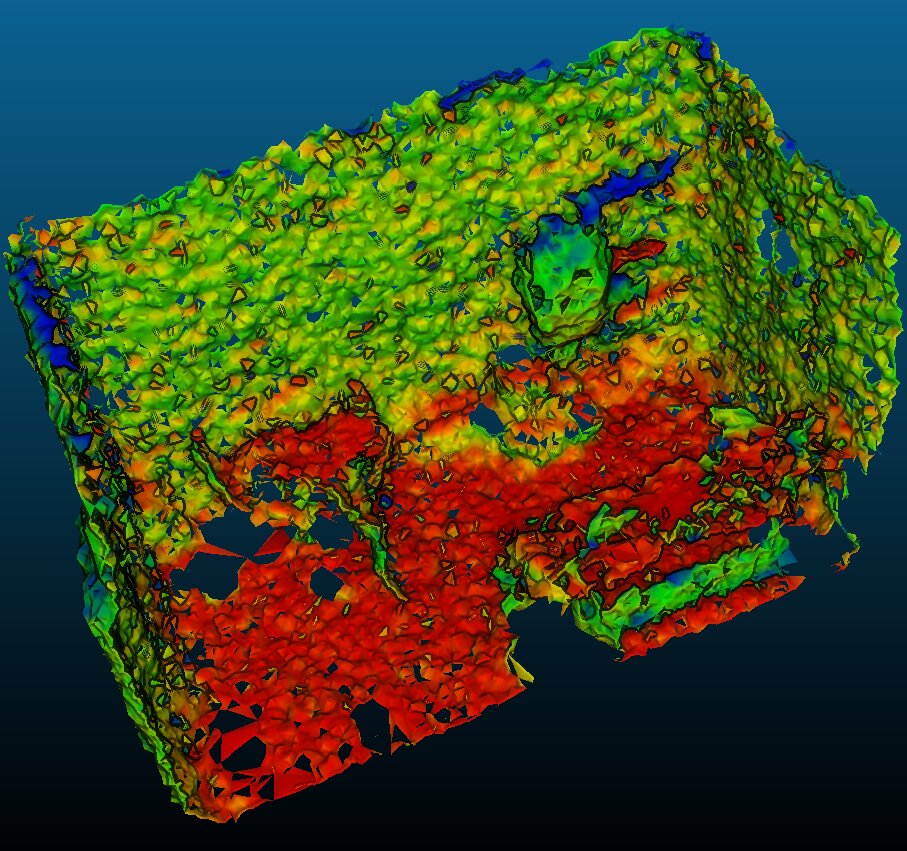}
        \caption{ToF Ball Pivoting Algorithm}
        \label{fig:tof_bpa}
    \end{subfigure}
    \hfill
    \begin{subfigure}[t]{0.49\linewidth}
        \centering
        \includegraphics[width=\linewidth,height=2.1in,keepaspectratio]{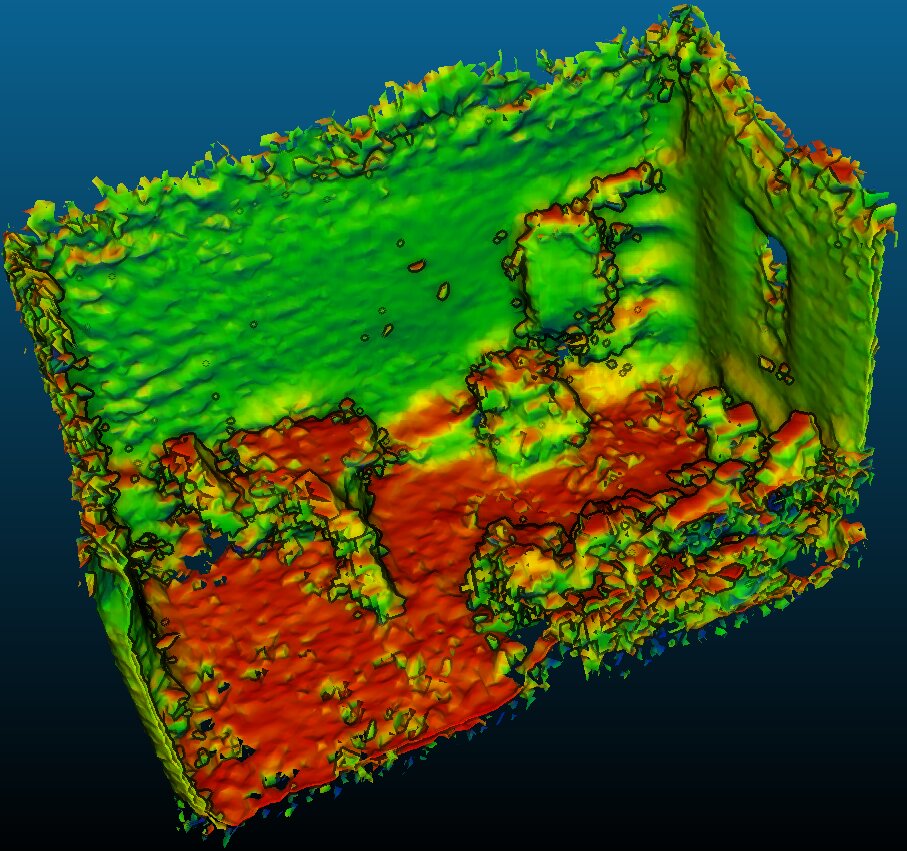}
        \caption{ToF TSDF volumetric fusion}
        \label{fig:tof_tsdf}
    \end{subfigure}

    \caption{Comparison between the ToF surface reconstruction and the LiDAR mesh using three meshing methods. All views are colorized by surface normal in the z-direction.}
    \label{fig:mesh_views}

    \vspace{0.5em}

    \captionof{table}{Surface-sampled point-to-surface distance metrics from the ToF mesh to the LiDAR Poisson reference mesh.}
    \label{tab:mesh_metrics_tof_to_lidar}

    % We make the font of the table smaller
    \footnotesize
\begin{tabular}{lccc}
\toprule
\multirow{2}{*}{\textbf{Metric}} & \multicolumn{3}{c}{\textbf{Reconstruction Method}} \\
\cmidrule(lr){2-4}
 & \textbf{ToF Poisson} & \textbf{ToF TSDF} & \textbf{ToF BPA} \\
\midrule
Mean (m)   & 0.028 & 0.071 & 0.068 \\
RMSE (m)   & 0.037 & 0.094 & 0.080 \\
Median (m) & 0.022 & 0.048 & 0.071 \\
P90 (m)    & 0.059 & 0.174 & 0.120 \\
P95 (m)    & 0.076 & 0.194 & 0.133 \\
Max (m)    & 0.245 & 0.342 & 0.237 \\
\bottomrule
\end{tabular}
\end{figure}

We evaluated the use of the distributed \gls{tof} sensors for different perception tasks. We first investigated their performance for 3D reconstruction (\secref{ssec:reconstruction}). Then, we integrated the system described in the previous section on the ANYbotics \anymal{} and evaluated footstep planning (\secref{ssec:footstep_planning}) and obstacle avoidance (\secref{ssec:obstacle_avoidance})\rev{, with pose provided throughout by the platform's default state estimator.}

%%%%%%%%%%%%%%%%%%%%%%%%%%%%%%%%%%%%%%%%%%%%%%%%%%%%%%%%%%%%%%%%%%%%%
% \subsection{Exp 1 -- Reconstruction Accuracy Against LiDAR sensors}\label{ssec:reconstruction}
\subsection{Exp 1 -- Reconstruction Accuracy \revv{with a LiDAR Reference}}\label{ssec:reconstruction}

% We first compared the performance of the ToF sensor arrays with a standard LiDAR device in mapping tasks, using the handheld configuration presented in \secref{ssec:handheld-integration}. 
We first \revv{evaluated the geometric reconstruction accuracy of the ToF sensor arrays, taking a standard LiDAR device as a reference}, using the handheld configuration presented in \secref{ssec:handheld-integration}. For this experiment, we focused on indoor scenes, e.g., office rooms.
% The mapping device was carried by hand and moved around to scan out full coverage of the room, with experiments conducted under standard indoor illumination.
\rev{The mapping device was carried by hand along a trajectory designed to sweep all room surfaces visible to both sensors, varying height and tilt to compensate for the ToF ring's narrower vertical FoV, under standard indoor illumination. A LiDAR-inertial odometry \revv{derived from VILENS~\cite{wisth2022vilens}, }% A LiDAR-inertial odometry, formulated as a factor-graph optimisation, 
provided the 6-DoF poses to project both the LiDAR and ToF clouds into a common reference frame for cross-evaluation}.
% We ran a \rev{a LiDAR-inertial state estimator, formulated as a factor-graph optimisation, to obtain accurate 6 DoF poses}, which we used to project both the LiDAR clouds and the ToF clouds into 3D. In this way we could acquire a LiDAR map and a ToF map of the room in the same reference frame for easy cross-evaluation of reconstruction accuracy. 

% We first evaluated the ToF point cloud map using bidirectional nearest-neighbor Euclidean distances between ToF and LiDAR point clouds. \figref{fig:pc_comparison_views} shows the ToF and LiDAR point clouds accumulated by scanning the office room (overall dimensions 4\,m $\times$ 5\,m).
\shrink{We evaluated the ToF point cloud map using bidirectional nearest-neighbor Euclidean distances to the LiDAR cloud (Fig. 5, office room, 4 m × 5 m).}
\rev{The LiDAR cloud is denser and more complete, partly due to its substantially wider vertical FoV, though the main room features remain identifiable in the ToF map.}
\shrink{Table~\ref{tab:tof_lidar_distance_metrics} reports these distances}\rev{; the larger maximum values are driven by distribution tails (median below 2 cm), reflecting incompletely observed or sparsely sampled regions\revv{, not} systematic reconstruction error.} 
% Table~\ref{tab:tof_lidar_distance_metrics} shows bidirectional nearest-neighbor distance metrics. \rev{The larger maximum values are driven by the tails of the distribution with median distances below 2 cm reflecting a small number of incompletely observed or sparsely sampled regions rather than a systematic reconstruction error.}

\shrink{Additionally, we evaluated whether the measurements support surface-mesh reconstruction using three methods: Poisson Surface Reconstruction~\cite{kazhdan2006}, Ball Pivoting Algorithm (BPA)~\cite{bernardini1999}, and TSDF volumetric fusion~\cite{curless1996}, \revv{adopting} the VDBFusion implementation~\cite{vizzo2022sensors}, meshing the LiDAR cloud with Poisson's method as a \rev{reference} for comparison}.
% Additionally, we further evaluated whether the measurements could support surface-mesh reconstruction using three methods: Poisson Surface Reconstruction~\cite{kazhdan2006}, Ball Pivoting Algorithm (BPA)~\cite{bernardini1999}, and TSDF volumetric fusion~\cite{curless1996}, using the VDBFusion implementation~\cite{vizzo2022sensors}. In this case we meshed the LiDAR point cloud using Poisson's method, which we used as \rev{a reference for comparison}.
% ground truth.

\shrink{\figref{fig:mesh_views} and Table~\ref{tab:mesh_metrics_tof_to_lidar} compare the ToF meshes against the LiDAR mesh: Poisson reconstruction is smoothest and most complete, BPA preserves local structure but leaves holes in sparsely sampled regions, and TSDF improves completeness via multi-view fusion at the cost of rougher, noisier surfaces. Overall, the ToF sensors reliably reconstruct the coarse geometry of indoor scenes, recovering structures down to approximately 10~cm in size.}
% \figref{fig:mesh_views} and Table~\ref{tab:mesh_metrics_tof_to_lidar} compare the resulting ToF meshes against the LiDAR mesh. Qualitatively, Poisson reconstruction produces the smoothest and most complete ToF mesh, BPA preserves local structure but introduces holes in sparsely sampled regions, and TSDF improves completeness through multi-view fusion but is more affected by ToF noise, leading to rougher surfaces and some spurious geometry. Overall, these results indicate that the ToF sensors are effective for reconstructing the coarse geometry of indoor scenes, reliably recovering structures down to approximately 10~cm in size.

%%%%%%%%%%%%%%%%%%%%%%%%%%%%%%%%%%%%%%%%%%%%%%%%%%%%%%%%%%%%%%%%%%%%%
\subsection{Exp 2 -- Footstep Planning} \label{ssec:footstep_planning} 

\shrink{Next, we evaluated whether the proposed ToF sensing system could support locomotion over challenging terrain, compared to standard depth-camera terrain mapping, testing ANYmal D on progressively more demanding traversal tasks. The objective was to determine whether an elevation map built solely from ToF geometry could provide sufficient near-field terrain information for the robot to detect raised obstacles and adapt its footholds.}
% Next, we evaluated whether the proposed ToF sensing system could support locomotion over challenging terrain, when compared to standard depth camera-based terrain mapping. We tested the \anymal{} quadruped on a sequence of progressively more demanding terrain traversal tasks. The objective was to determine whether an elevation map generated solely from the distributed ToF sensors could provide sufficient near-field terrain information for the robot to detect raised obstacles and adapt its footholds. 

To demonstrate this, we tested the locomotion behavior under \rev{two} perception setups: (1) terrain maps generated from the distributed ToF sensors mounted on the front shell, and (2) terrain maps from the two front-mounted Intel RealSense D435i cameras already installed on \anymal{}. \rev{As a qualitative baseline, we also included (3) a blind configuration with no terrain-map input, using the default low-level controller (not fine-tuned), to confirm that exteroceptive feedback is necessary under these controller settings.}

We tested these configurations on three terrain courses: flat terrain with railway sleepers, an irregular indoor staircase, and an outdoor staircase.

\begin{figure}[t]
    \centering
    \captionsetup{font=small}
    % \begin{subfigure}[t]{\linewidth}
    %     \centering
    %     \includegraphics[width=\linewidth]{figures/fig-sleepers-blur-wide.png}
    %     \label{fig:sleepers}
    % \end{subfigure}
    % % \vspace{0.5em}

    % \begin{subfigure}[t]{0.475\linewidth}
    %     \centering
    %     \includegraphics[width=\linewidth]{figures/fig-zoom-depth-sleepers.png}
    %     \label{fig:depth_sleepers}
    % \end{subfigure}
    % \hfill
    % \begin{subfigure}[t]{0.475\linewidth}
    %     \centering
    %     \includegraphics[width=\linewidth]{figures/fig-zoom-tof-sleepers.png}
    %     \label{fig:tof_sleepers}
    % \end{subfigure}

    \includegraphics[width=.9\linewidth]{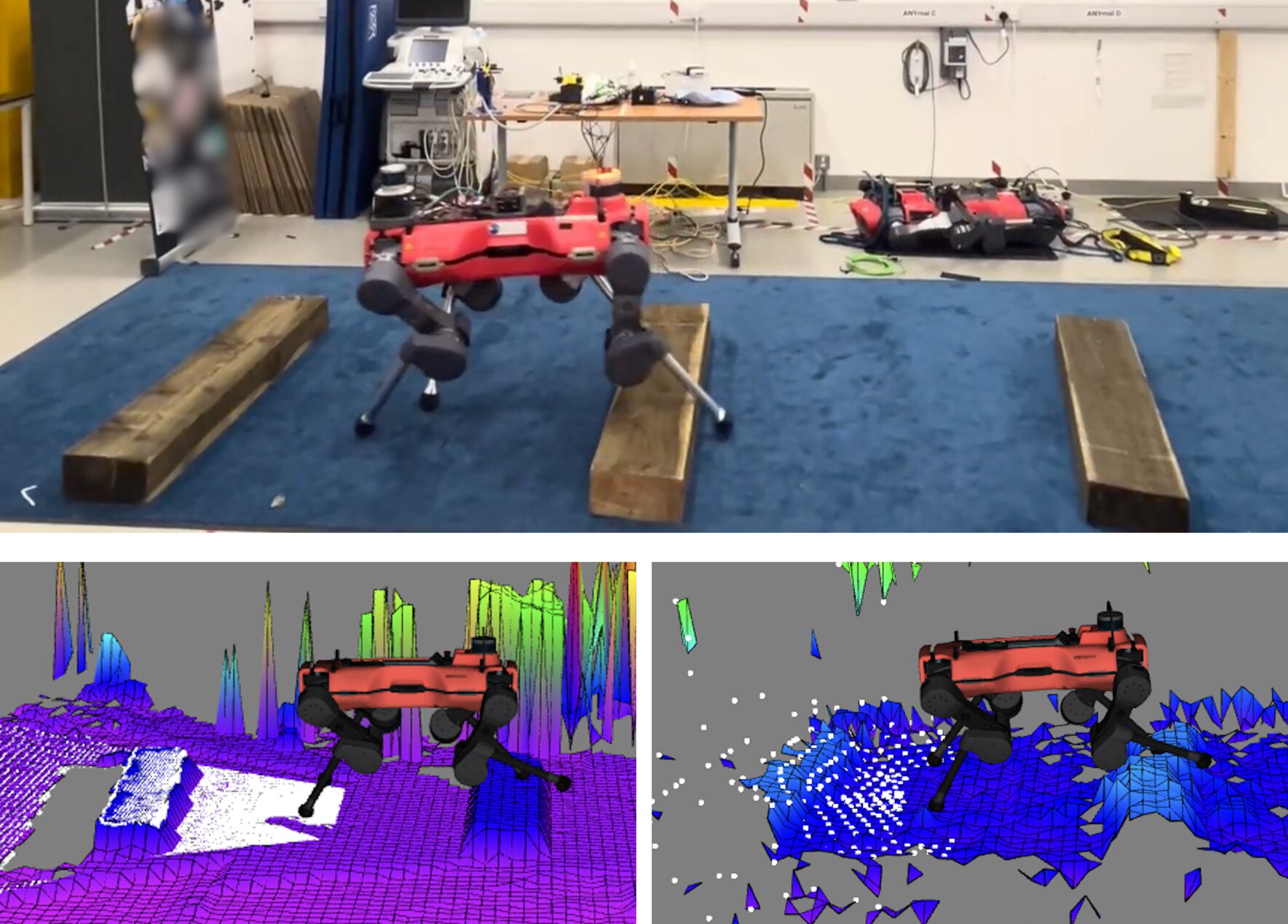}

    \caption{Exp 2 -- Footstep Planning on Railway Sleepers. Top: \anymal{} traversing an obstacle course consisting of three 13~cm railway sleepers arranged in line. Bottom: representative elevation maps generated using depth sensing (left) and distributed ToF sensing (right).}
    \label{fig:footstep_course}
\end{figure}

% \noindent\textbf{Railway Sleepers.}\label{ssec:railway-sleepers}
% First, we teleoperated the \anymal{} robot to traverse in 10 trials a course of three railway sleepers arranged along the robot's direction of travel (see \figref{fig:footstep_course}), where each sleeper was about 13\,cm high. Across the trials, for each perception setup, the terrain map built from each perceptive modality was fed to a perceptive reinforcement learning-based locomotion controller~\cite{miki2022learning}, enabling obstacle-aware footstep adaptation. Trials were considered successful if the robot crossed the entire course, stepping over each sleeper without tripping or falling.
\noindent\textbf{Railway Sleepers.}\label{ssec:railway-sleepers} 
We evaluated locomotion over a course of three railway sleepers arranged along 
the robot's direction of travel (see \figref{fig:footstep_course}), each 
approximately 13\,cm high. The robot was commanded by an operator providing 
high-level forward velocity commands, while footstep planning were handled 
autonomously by a perceptive reinforcement learning-based locomotion 
controller~\cite{miki2022learning} using the terrain map built from each modality. We ran 10 trials per perception setup. Trials were considered 
successful if the robot crossed the entire course without tripping or falling.

Using the ToF-based and depth-camera-based elevation map, the robot completed all the trials successfully, at an average commanded velocity of 40\,cm/s.
% (see \tabref{tab:traversal_performance}). 
\rev{Without \revv{terrain-perception feedback}, it failed in all 3 attempts, each starting from a slightly different initial position.}

\begin{comment}
\begin{table}[!t]
\centering
\captionsetup{font=small}
\caption{Performance on the railway-sleeper, indoor staircase, and outdoor staircase traversal tasks, repeated over 10 trials, under ToF-based elevation mapping, depth-camera-based elevation mapping, and blind operation.}
\label{tab:traversal_performance}
\footnotesize
\setlength{\tabcolsep}{3pt}
\begin{tabular}{p{2cm} p{3cm} p{1.5cm} p{1.2cm}}
\toprule
\textbf{Task} & \textbf{Configuration} & \textbf{Avg. Vel.} & \textbf{Success} \\
              &                        & \textbf{[cm/s]}    & \textbf{[\%]} \\
\midrule
Railway sleepers
& Depth Cameras Elev.\ Map & 40 & 100 \\
& ToF Elev.\ Map           & 40 & 100 \\
& Blind                         & -- & 0 \\
\midrule
Indoor staircase
& Depth Cameras Elev.\ Map & 15 & 100 \\
& ToF Elev.\ Map           & 15 & 100 \\
& Blind                         & -- & 0 \\
\midrule
Outdoor staircase
& Depth Cameras Elev.\ Map & 25 & 100 \\
& ToF Elev.\ Map           & 25 & 100 \\
& Blind                         & -- & 0 \\
\bottomrule
\end{tabular}
\end{table}
\end{comment}

%

\begin{figure}[t]
    \centering
    \captionsetup{font=small}
    % \begin{subfigure}[t]{0.9\linewidth}
    %     \centering
    %     \includegraphics[width=\linewidth]{figures/fig-indoorstaircase-wide.png}
    %     \label{fig:indoor_staircase_photo}
    % \end{subfigure}

    % \vspace{0.5em}

    % \begin{subfigure}[t]{0.475\linewidth}
    %     \centering
    %     \includegraphics[width=\linewidth,height=1.8in,keepaspectratio]{figures/fig-zoom-depth-indoor.png}
    %     \label{fig:depth_indoor_staircase}
    % \end{subfigure}
    % \hfill
    % \begin{subfigure}[t]{0.475\linewidth}
    %     \centering
    %     \includegraphics[width=\linewidth,height=1.8in,keepaspectratio]{figures/fig-zoom-tof-indoor.png}
    %     \label{fig:tof_indoor_staircase}
    % \end{subfigure}

 \includegraphics[width=0.9\linewidth]{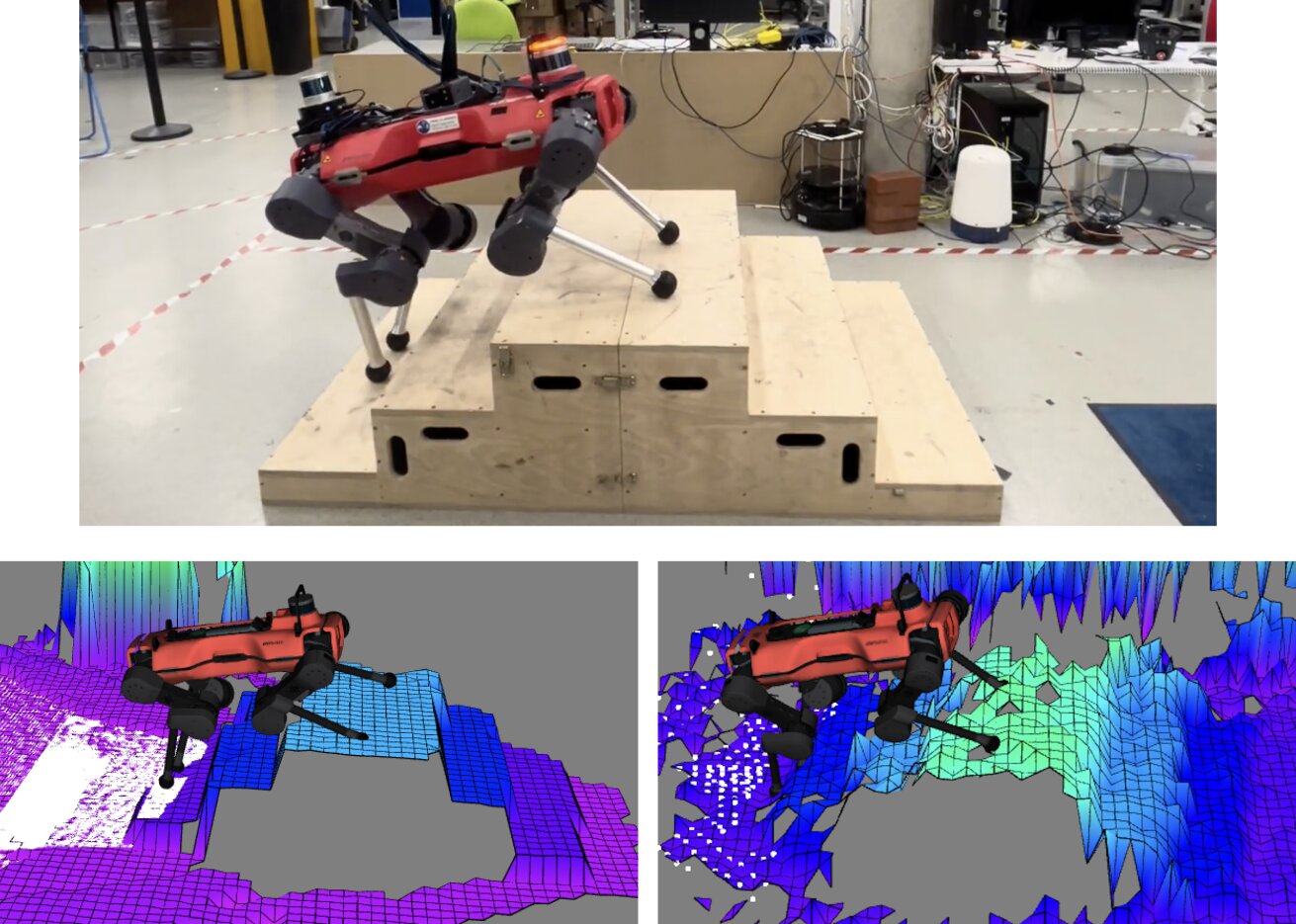}

    \caption{Exp 2 -- Footstep Planning on an Irregular Indoor Staircase. Top: indoor staircase experiment with \anymal{}. Bottom: representative elevation maps generated using depth sensing (left) and distributed ToF sensing (right).}
    \label{fig:indoor_staircase}
\end{figure}

\noindent\textbf{Irregular Indoor Staircase.}
We next evaluated indoor staircase climbing
% . We used 
\shrink{on} a custom staircase with irregular steps, with heights of 9\,cm, 17\,cm, and 17\,cm
% , as shown in  
\shrink{(\figref{fig:indoor_staircase}), teleoperating the robot to ascend and descend 10 times.}

\shrink{The depth-camera map clearly showed the step discontinuities, while the noisier ToF map still preserved a distinguishable staircase structure, sufficient for the RL controller to adapt its footholds. Both configurations succeeded in all 10 attempts; in the blind condition, the robot typically cleared the first 9 cm step but collided with the subsequent steps.}
%. We followed the same protocol, teleoperating the robot to ascend and descend the staircase 10 times.

% The depth-camera configuration produced a clean elevation map in which the step discontinuities were clearly visible. Although the ToF-based elevation map contained noise and local artifacts, the overall staircase structure remained distinguishable. This was sufficient for the reinforcement-learning locomotion controller to adapt its footholds as required. Both the depth-camera and ToF sensing configurations enabled the successful traversing of the staircase in all 10 attempts.
% (\tabref{tab:traversal_performance}). 
% In blind operation, the robot was typically able to clear the first 9\,cm step, but then collided with the higher subsequent steps and failed to complete the staircase.

\begin{figure}[t]
    \centering
    \captionsetup{font=small}
    % \begin{subfigure}[t]{0.475\linewidth}
    %     \centering
    %     \includegraphics[width=\linewidth,height=1.17in]{figures/stairs_readjusted.jpeg}
    %     \label{fig:outdoor_staircase_extra}
    % \end{subfigure}
    % \hfill
    % \begin{subfigure}[t]{0.475\linewidth}
    %     \centering
    %     \includegraphics[width=\linewidth,height=1.4in,keepaspectratio]{figures/fig-outdoorstaircase.png}
    %     \label{fig:outdoor_staircase_photo}
    % \end{subfigure}

    % \vspace{0.5em}

    % \begin{subfigure}[t]{0.475\linewidth}
    %     \centering
    %     \includegraphics[width=\linewidth,height=1.8in,keepaspectratio]{figures/fig-zoom-depth-outdoor.png}
    %     \label{fig:depth_outdoor_staircase}
    % \end{subfigure}
    % \hfill
    % \begin{subfigure}[t]{0.475\linewidth}
    %     \centering
    %     \includegraphics[width=\linewidth,height=1.8in,keepaspectratio]{figures/fig-zoom-tof-outdoor.png}
    %     \label{fig:tof_outdoor_staircase}
    % \end{subfigure}

    \includegraphics[width=0.9\linewidth]{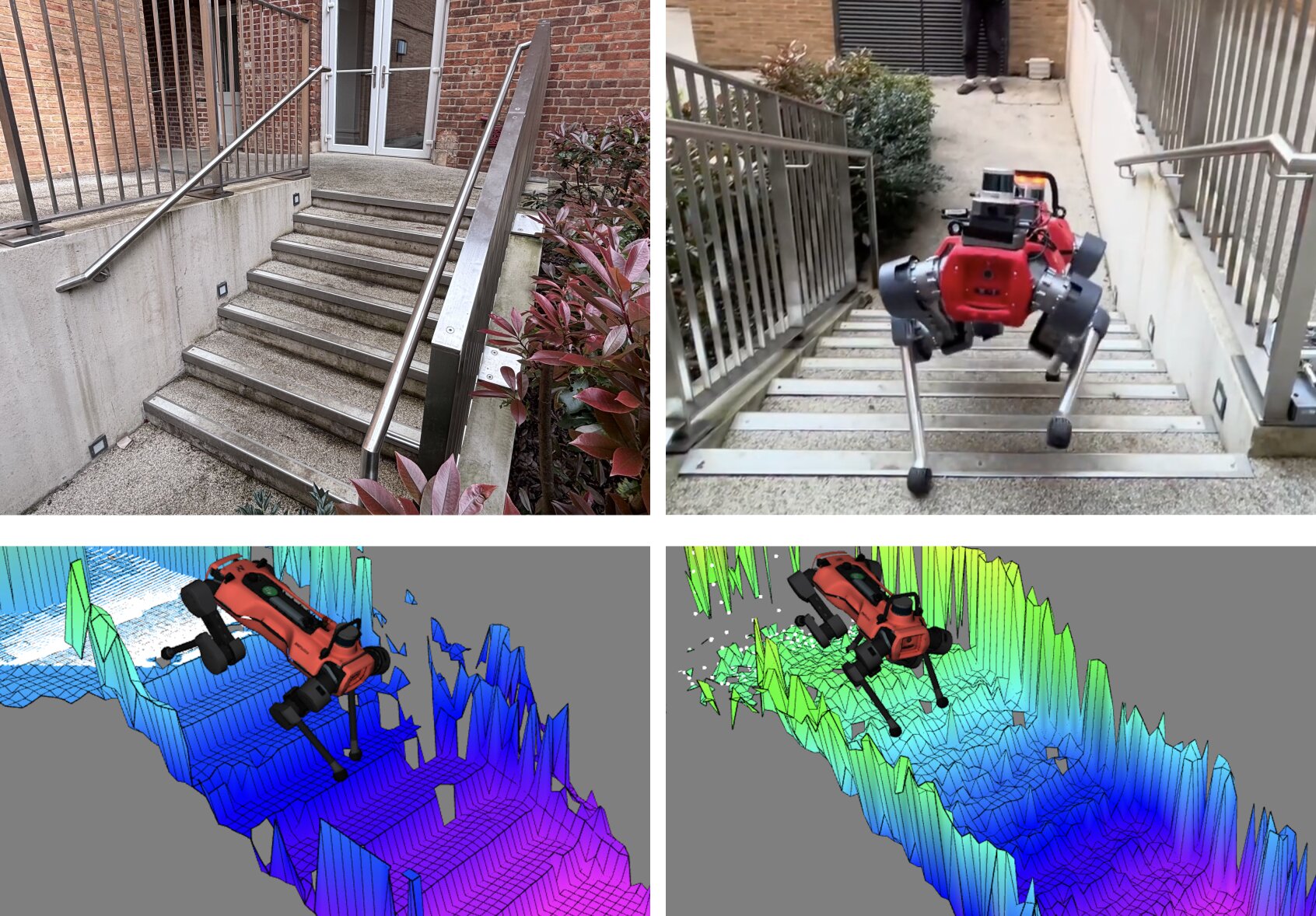}
    \caption{Exp 2 -- Footstep Planning on an Outdoor Staircase. Top: \anymal{} during the outdoor staircase experiment. Bottom: representative elevation maps generated using depth sensing (left) and distributed ToF sensing (right).}
    \label{fig:outdoor_staircase}
\end{figure}

\noindent\textbf{Outdoor Staircase.}
\revv{Lastly, we tested the platform on an outdoor staircase of seven 15\,cm steps under the same protocol} (see \figref{fig:outdoor_staircase}). \revv{Results} matched the indoor experiments
% Lastly, we tested the platform on an outdoor staircase under the same protocol, comprising seven steps, each of 15\,cm height (see \figref{fig:outdoor_staircase}). The results were consistent with the indoor experiments
\shrink{: the depth camera produced the highest-fidelity maps, while the coarser ToF map still captured the stepped structure well enough for ANYmal D to ascend despite its noise and artifacts. Both configurations succeeded over 10 trials; in the blind condition, the robot \revv{failed to climb the first step in all 3 attempts.}}
% could not climb the first step in \revv{any} of 3 attempts.}
% , where the depth camera produced the highest fidelity terrain maps. The ToF-based elevation map does not recover the stair geometry as sharply, but still captured the overall stepped structure sufficiently well for \anymal{} to ascend the staircase despite the map's noise and local artifacts. Over 10 repeated trials, both sensing configurations enabled successful traversal.
% In contrast, in blind operation, the robot was unable to climb the first step, across 3 attempts from different starting positions.

%

Together, these experiments support our claim that the proposed ToF sensing system is suitable for near-field terrain perception and locomotion over challenging stepped terrain, with similar performance to the depth cameras \shrink{(further documented in the supplementary video).}
% . These experiments are further documented in the supplementary video.

%%%%%%%%%%%%%%%%%%%%%%%%%%%%%%%%%%%%%%%%%%%%%%%%%%%%%%%%%%%%%%%%%%%%%
\subsection{Exp 3 -- Obstacle Avoidance} \label{ssec:obstacle_avoidance}

\begin{figure}[!t]
    \centering
    \captionsetup{font=small}
    \includegraphics[width=0.85\linewidth]{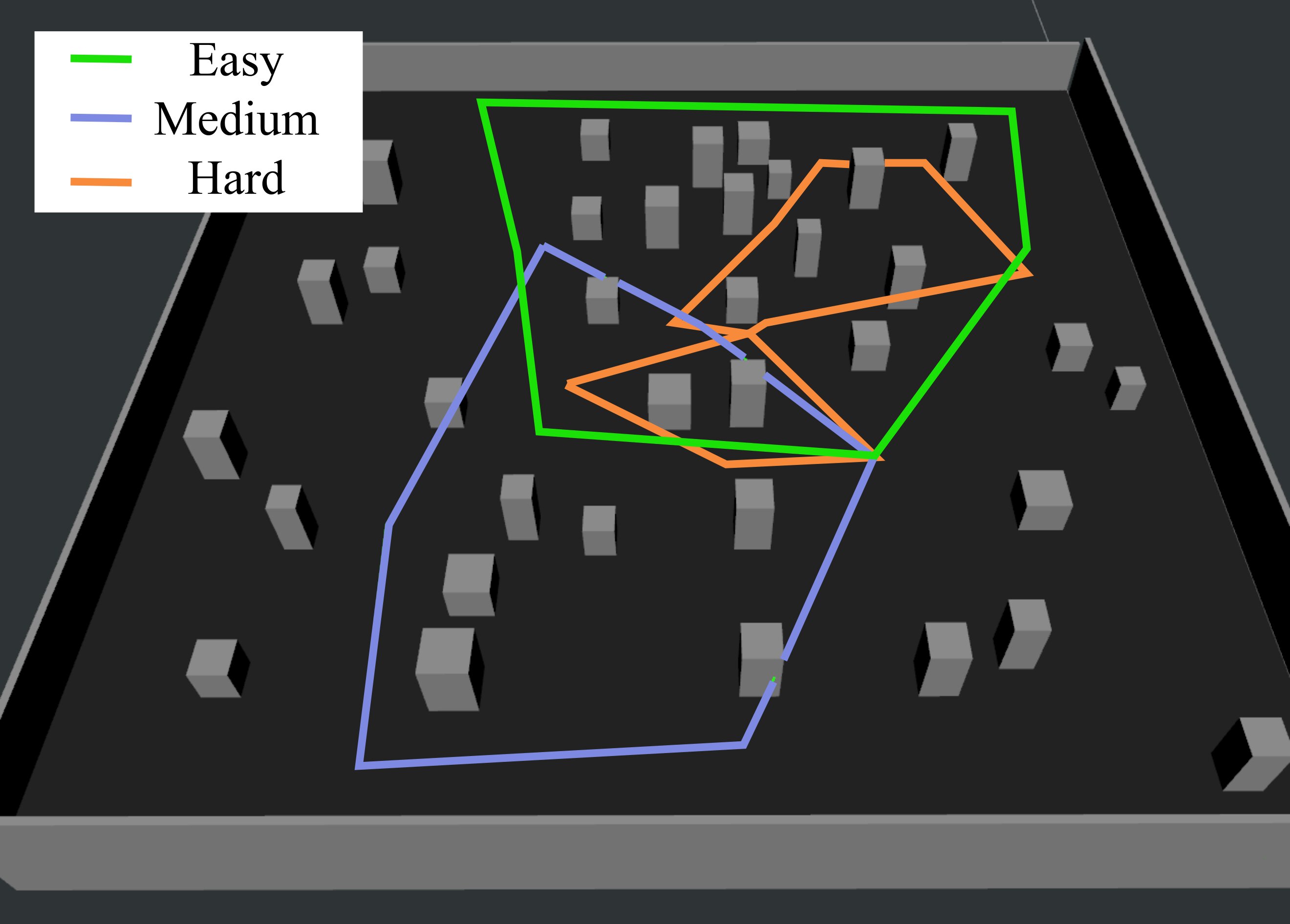}
    \caption{Simulation environment with navigation paths of increasing complexity used for obstacle avoidance.
    }
    \label{paths}
\end{figure}

\begin{figure*}[!t]
    \centering
    \captionsetup{font=small}
    \includegraphics[width=0.95\linewidth]{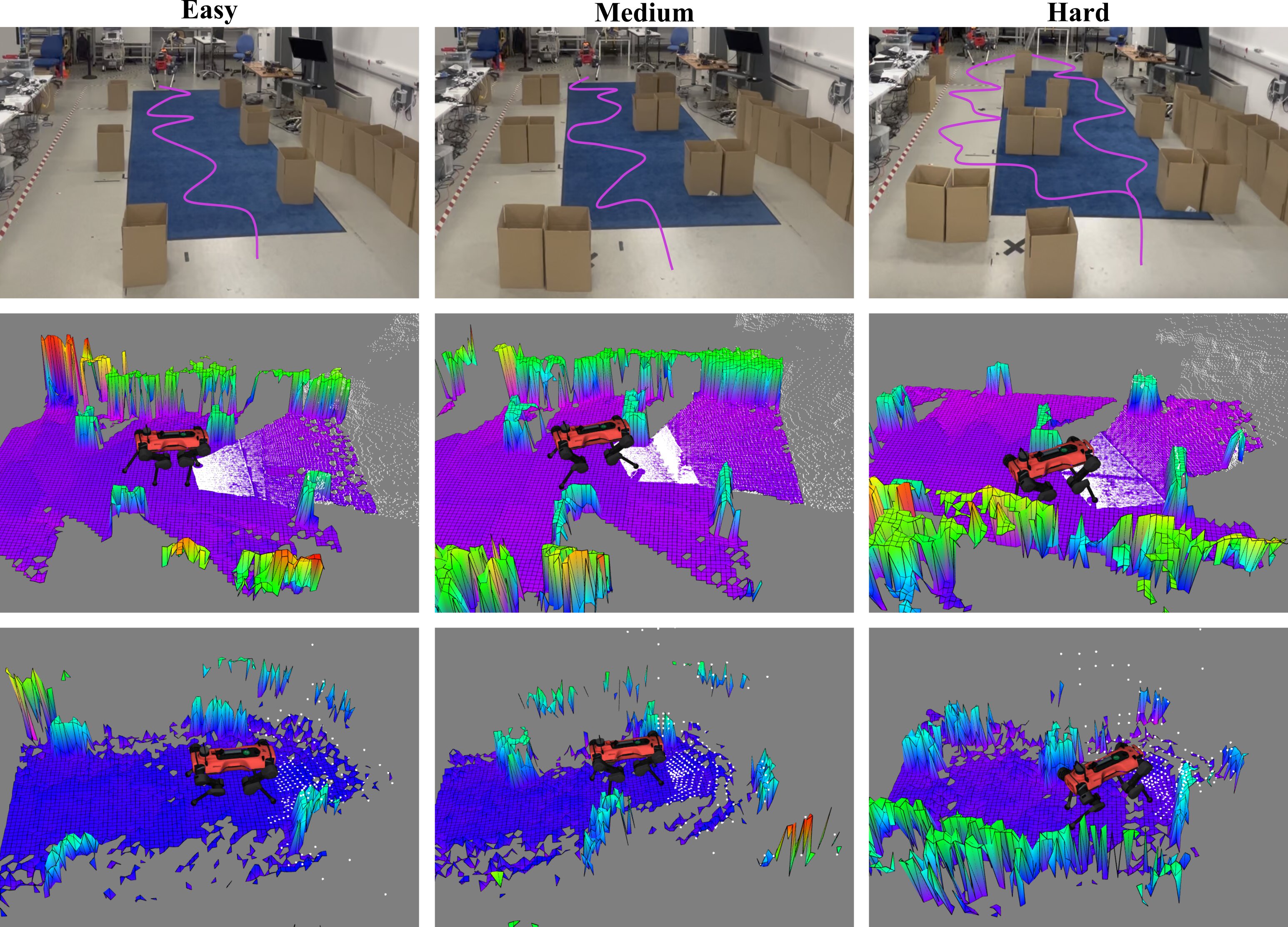}
    \caption{Real world obstacle-avoidance experiments under three difficulty levels. Columns correspond to easy, medium, and hard obstacle-course layouts. The top row shows the obstacle layouts together with the robot trajectories, the middle row shows the corresponding depth-camera-based perception, and the bottom row shows the corresponding ToF-based perception.}
    \label{fig:obstacle_avoidance_comparison}
\end{figure*}

Our final experiment aimed to determine whether a terrain map generated solely from the distributed ToF sensors could support autonomous local navigation through a cluttered obstacle course.
We first conducted preliminary experiments in simulation and then proceeded with a real-world deployment on the \anymal{} platform. In both cases, the robot navigated through the obstacle course using a reactive local planner \cite{mattamala2022ral} which uses the elevation maps built from either the distributed \gls{tof} sensors or the depth cameras as input.

\noindent\textbf{Simulation Experiments.}
We simulated the \gls{tof} sensors by combining the depth camera module already available in Gazebo with the noise model proposed by Caroleo et al. \cite{caroleo2025iros}. In particular, we introduced a linearly-scaled Gaussian noise model that increases with distance to replicate the variability of \gls{tof}s: $d_{\text{sim}} = d_{\text{GT}} + \mathcal{N}\left(0, \left(0.01\, d_{\text{GT}}\right)^2\right)$, where $d_{\text{sim}}$ is the simulated distance from the sensor to the detected point, while $d_{\text{GT}}$ is the ground truth distance. The simulated \gls{tof} array operates at coarser spatial resolution than the depth camera, reflecting the behaviour of the physical sensor.

For testing, we defined the simulated obstacle course shown in \figref{paths}. We designed 3 navigation routes to have increasing difficulty: the simplest one without obstacles and with wide passages and turns, to more complex paths with constrained sections, sharp turns and obstacles placed intersecting with the reference path. The robot was commanded to follow each course 10 times using the local planner.

Table~\ref{tab:navig} \revv{reports the local planner's performance, averaged over the 10 runs}
% reports the performance of the local planner (averaged over the 10 runs),
 \shrink{for both modalities,} % using either ToF sensors or depth cameras for elevation mapping, 
according to completion time and the distance traveled. All trials were successfully completed across each of the course difficulties, \rev{with path lengths close throughout; completion times are comparable at low difficulty but somewhat slower for ToF at medium and hard, as obstacle boundaries enter the coarser, shorter-range ToF map later, inducing more conservative local replanning.} 
% with comparable completion times and traveled distances. The elevation map progressively refines the reconstruction as the robot approaches obstacles in a manner which reduces noise from the \gls{tof} sensors, and improves local reconstruction.

\rev{Despite their coarser resolution, ToF sensors proved sufficient for detecting all relevant obstacles in this simulated task, with no marked advantage from the depth camera's higher resolution.}
% Despite their coarser spatial resolution, \gls{tof} sensors proved sufficient for detecting all relevant obstacles, rendering the higher resolution of depth cameras redundant for this task.
% The low resolution allowed the detection of relevant obstacles, making the higher resolution of depth cameras largely redundant.

%\mfallon{Matteo: write details about how you simulated the ToF sensor - saying that you repurposed the depth camera module in Gazebo.}

\begin{table}[b]
\captionsetup{font=small}
\centering
\caption{Performance of obstacle avoidance task for both depth camera and ToF configurations.}
\label{tab:navig}
\begin{tabular}{lcccc}
\toprule
\multirow{2}{*}{\textbf{Course}} & \multicolumn{2}{c}{\textbf{Completion Time [s]}} & \multicolumn{2}{c}{\textbf{Path Length [m]}} \\
\cmidrule(lr){2-3} \cmidrule(lr){4-5}
 & \textbf{Depth} & \textbf{ToF} & \textbf{Depth} & \textbf{ToF} \\
\midrule
Low difficulty    & 63.8 & 63.2 & 35.0 & 34.2 \\
Medium difficulty & 47.5 & 55.4 & 27.2 & 28.8 \\
Hard difficulty   & 71.8 & 84.6 & 28.5 & 30.5 \\
\bottomrule
\end{tabular}
\end{table}

\noindent\textbf{Fault Tolerance in Simulation.}
% The simulation environment was also used to evaluate the robustness of the system to possible sensor failures. Specifically, the objective was to determine whether, and to what extent, the proposed distributed system remains operational under such conditions or becomes incapable of functioning. Starting from the nominal configuration of 8 fully operational ToF sensors, we modified the simulator to reduce the set of sensors from eight to five to simulate failed ToF sensors. The sensor removal was randomized in each experiment to avoid bias induced by specific spatial sensor arrangements. Each configuration was tested 10 times for each of the three courses
\shrink{The simulation environment was also used to evaluate robustness to sensor failures, i.e., whether and how far the distributed system degrades gracefully. Starting from 8 fully operational sensors, we reduced the set to five, randomizing which sensors were removed in each trial to avoid bias from specific spatial arrangements. Each configuration was tested 10 times per course,}
% \mfallon{REVISED THIS: 
resulting in a total of 120 trials.
% }.

% \mfallon{There were no failed sensors - merely we removed a sensor. So there were no faulty range measurements or outliers... saying you REMOVED a sensor is clearer}

\shrink{Table~\ref{tab:redundancy} shows the system is robust to these `sensor failures'.}
% The results in Table~\ref{tab:redundancy} show that the system is robust to these `sensor failures'. 
\revv{Performance remains largely unaffected when two sensors are removed, with success rates close to nominal; a noticeable degradation appears only when three are removed, particularly on the more complex obstacle course.}
% Performance remains largely unaffected when two sensors were removed, maintaining success rates close to nominal. A noticeable degradation appeared when three sensors were removed, particularly in the more complex obstacle course.

% Overall, the system tolerates up to two sensor failures with minimal impact, while performance degrades significantly beyond this point.

\begin{table}[b]
\captionsetup{font=small}
\centering
\caption{Success rate (\%) of simulated obstacle avoidance when progressively removing ToF sensors (from a starting point of 8).}
\label{tab:redundancy}
\begin{tabular}{lcccc}
\toprule
\multirow{2}{*}{\textbf{Course}} & \multicolumn{4}{c}{\textbf{Number of Sensors Remaining}} \\
\cmidrule(lr){2-5}
 & \textbf{8} & \textbf{7} & \textbf{6} & \textbf{5} \\
\midrule
Low difficulty    & 100 & 100 & 100 & 92.5 \\
Medium difficulty & 100 & 97.5 & 97.5 & 85.0 \\
Hard difficulty   & 100 & 95.0 & 95.0 & 67.5 \\
\bottomrule
\end{tabular}
\end{table}

\begin{table}[!t]
\centering
\captionsetup{font=small}
\caption{Obstacle-avoidance performance across three difficulty levels, repeated over 10 trials. Completion time is averaged over successful trials only. We report the average number of collisions of the \anymal{}'s legs with an obstacle for runs that still reached the goal. Failures are reported out of 10 trials.}
\label{tab:rw-obstacle-avoidance}
\footnotesize
\setlength{\tabcolsep}{3.5pt}
\begin{tabular}{lccc ccc}
\toprule
\multirow{2}{*}{\textbf{Course}} & \multicolumn{3}{c}{\textbf{Depth}} & \multicolumn{3}{c}{\textbf{ToF}} \\
\cmidrule(lr){2-4} \cmidrule(lr){5-7}
 & \textbf{Time [s]} & \textbf{Collisions} & \textbf{Fail.} & \textbf{Time [s]} & \textbf{Collisions} & \textbf{Fail.} \\
\midrule
Easy    & 29.7 & 0.0 & 0\% & 27.5 & 0.2 & 0\% \\
Medium & 32.9 & 0.1 & 0\% & 37.0 & 0.56 & 10\% \\
Hard   & 66.1 & 1.5 & 20\% & 71.4 & 1.63 & 20\% \\
\bottomrule
\end{tabular}
\end{table}

\noindent\textbf{Obstacle Avoidance in Real-World.}
To evaluate whether the proposed sensing setup could support fully autonomous local navigation, we conducted real-world obstacle-avoidance experiments using cardboard boxes arranged into three cluttered layouts of increasing difficulty, denoted as easy, medium, and hard, as shown in \figref{fig:obstacle_avoidance_comparison}. In each trial, \anymal{} was assigned a goal position beyond the obstacle field and navigated using an online local elevation map. As the robot moved, new sensor observations were fused into the map, and the local planner generated reactive motion plans toward the goal. We compared two perception conditions: elevation maps generated from the front-mounted depth cameras and from the distributed ToF sensors.

Table~\ref{tab:rw-obstacle-avoidance} summarizes the results over 10 trials per difficulty level. \rev{A trial was considered successful if the robot reached the goal; failures corresponded to cases where the local planner could not make progress toward the goal.} The depth-camera configuration produced cleaner and more spatially complete local maps, resulting in fewer contacts and slightly more reliable execution. The ToF configuration was still sufficient to capture the dominant obstacle structure and support goal-reaching navigation, but its sparser sensing and less complete geometric reconstruction of the cardboard boxes led to more frequent leg contacts, particularly as the robot moved around obstacle boundaries. Failures in both configurations were partly associated with local-minima behavior of the reactive planner in tightly constrained layouts~\cite{mattamala2022ral}, where the attractive field toward the local goal and the repulsive fields around nearby obstacles canceled each other, causing the robot to become stuck in its configuration and not make progress toward the end goal.
\rev{Some ToF failures were also caused by sparse reconstruction of nearby objects, such as the table on the course's right-hand side, leading to incomplete obstacle boundaries in the local map.}

Overall, these trials show that the distributed ToF array provides sufficient geometric information for practical local navigation on \anymal{}. Although the ToF-based elevation maps are lower in fidelity than \revv{the depth cameras'}, they still allow the robot to detect the main obstacle structure, plan feasible motions, and reach the goal reliably across cluttered indoor layouts. The main limitations lie in reduced reconstruction precision \rev{and sparser angular resolution}, which \revv{most} affect leg clearance in tight maneuvers \rev{and the detection of thin or small obstacles}, but the results nevertheless support distributed ToF sensing as a viable lightweight modality for conservative near-field obstacle avoidance. \rev{This aligns with the scope of our approach: the robot-centric elevation map is \revv{rebuilt continually rather than kept globally consistent}, suiting short-horizon tasks like footstep planning and near-field obstacle avoidance.
% despite the ToF array's coarser resolution.
For long-range navigation, where sustained accuracy beyond the sensors' immediate view is required, the ToF reconstruction could complement higher-fidelity exteroceptive sensing.}

%% file: sections/05_conclusions.tex
\section{Conclusions}\label{sec:conclusions}
This work investigated the use of distributed Time-of-Flight (ToF) sensors as a low-cost and energy-efficient perception solution for quadruped locomotion and navigation. \rev{Our results show that distributed ToF sensing provides sufficient near-field terrain information for reliable footstep planning and local autonomous navigation through cluttered scenarios}\revv{, at substantially lower cost, power consumption, and system complexity than depth cameras, despite the coarser and less complete maps.}

Limitations remain in sensing resolution and noise sensitivity\revv{; distributed ToF sensing therefore complements rather than replaces the higher-fidelity sensors required for persistent mapping and long-range navigation.} Future work will explore multi-modal sensor fusion and varying the sensor distribution to improve robustness.
% This work investigated the use of distributed Time-of-Flight (ToF) sensors as a low-cost and energy-efficient perception solution for quadruped locomotion and navigation. \rev{Our results show that distributed ToF sensing provides sufficient near-field terrain information for reliable footstep planning and local autonomous navigation through cluttered scenarios, offering a competitive alternative to depth cameras in terms of cost, power consumption, and system complexity.} 
% % Our results show that distributed ToF sensing provides sufficient near-field perception for reliable footstep planning and autonomous navigation through cluttered spaces, achieving performance comparable to depth cameras with reduced cost, power consumption, and system complexity.

% Limitations remain in sensing resolution and noise sensitivity. Future work will explore multi-modal sensor fusion and varying the sensor distribution to improve robustness.